\documentclass{article}

\usepackage{iclr2027_conference,times}
\iclrfinalcopy
\usepackage[T1]{fontenc}
\usepackage[utf8]{inputenc}
\usepackage{lmodern}
\usepackage{textcomp}

\usepackage{amsmath,amssymb}
\usepackage{graphicx}
\usepackage{booktabs,array,longtable}
\usepackage{placeins}
\usepackage{wrapfig}

\usepackage{xcolor}
\usepackage{xspace}
\usepackage[hidelinks]{hyperref}
\usepackage{url}

\graphicspath{{figures/}}
\newcolumntype{L}[1]{>{\raggedright\arraybackslash}p{#1}}

\newcommand{\stagq}{StagQ\xspace}
\newcommand{\stagxs}{StagQ-XS\xspace}
\newcommand{\stagxxs}{StagQ-XXS\xspace}
\newcommand{\bpw}{\ensuremath{\mathrm{bpw}}\xspace}

\DeclareUnicodeCharacter{2212}{\ensuremath{-}}
\DeclareUnicodeCharacter{00D7}{\ensuremath{\times}}
\DeclareUnicodeCharacter{2248}{\ensuremath{\approx}}
\DeclareUnicodeCharacter{2264}{\ensuremath{\le}}
\DeclareUnicodeCharacter{2265}{\ensuremath{\ge}}
\DeclareUnicodeCharacter{00B1}{\ensuremath{\pm}}
\DeclareUnicodeCharacter{2192}{\ensuremath{\rightarrow}}

\begin{document}

\title{\stagq: Constraint-Driven Multi-Precision Weight Quantization for LLMs}
\author{{\spaceskip=0.8\fontdimen2\font Zhe Wei, Mengqi Guo, Yuan Yuan, Jiunn Bin Lim, Boyi Pan, Michael Bi Mi} \\
Huawei Technologies Ltd.}
\date{}
\maketitle
\lhead{Preprint}

\begin{abstract}
Serving a large language model (LLM) across a fleet of deployments
requires several weight-precision operating points.  Multi-precision formats serve them all from one stream whose prefixes
are valid lower-precision codes, instead of storing multiple copies.  We present
\stagq, a multi-precision weight format whose main stream is a 2-bit
group-wise affine base followed by a configurable number of 1-bit refinement planes on a
dyadic step schedule.  Every supported precision is a readable
prefix, decoded by an affine map derived from metadata shared across all precisions, with no per-weight lookup.  A sparse side record, filled both
before and after the grid is fitted, holds out the few weights the grid
serves worst.  We report two
configurations of the encoder.  At two bits the cheaper one leads the strongest multi-precision
baseline on Llama-3.1-8B, Phi-4 and OLMo-2-7B by \(3.1\) to \(7.0\) MMLU points, at a slightly lower logical rate.
At three bits it leads on Llama-3.1-8B, leads on Phi-4 at a higher
rate, and ties on OLMo-2-7B.  At four bits it ties on all three, at a
higher rate.  In a batch-one matrix--vector product on an NVIDIA A100 GPU,
timed on synthetic weights, our kernel is faster than the two baseline kernels in most
shape--precision cases.
\end{abstract}

\section{Introduction}
\label{sec:introduction}

Serving a large language model (LLM) is increasingly constrained by
memory capacity and memory traffic.  In low-batch decoding every weight
is read once per generated token, and decode speed is usually limited
by those reads rather than by arithmetic.  Weight-only post-training
quantization (PTQ) is the established response.  A deployment, however,
rarely runs at a single operating point.  Capacity, concurrency,
latency targets and request priority vary across a fleet.  With conventional PTQ~\citep{gptq, awq}, each precision is a separate
checkpoint that has to be quantized, validated and stored on its own.

\begin{wrapfigure}{r}{0.45\linewidth}
  \vspace{-\intextsep}
  \centering
  \includegraphics[width=\linewidth]{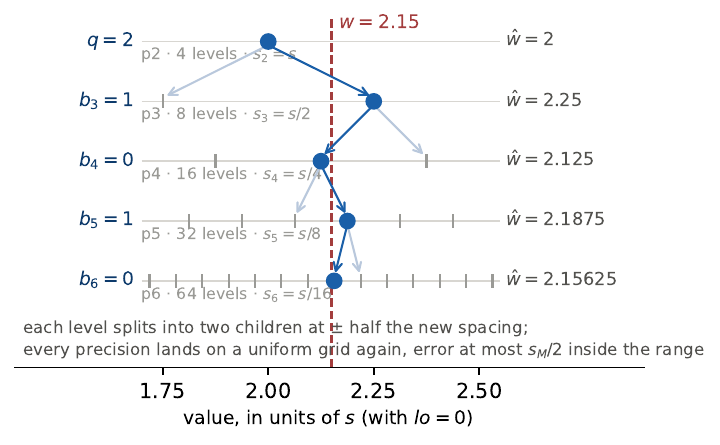}
  \caption{The \stagq main stream.  Each added plane halves the spacing,
  and every precision is again a uniform grid.  The weight \(w=2.15\) is encoded as \(q{=}2,\,b_3{=}1,\,b_4{=}0,\,b_5{=}1,\,b_6{=}0\).}
  \label{fig:ladder-example}
\end{wrapfigure}

A multi-precision format stores the weights once, as a stream in which
every prefix is itself a valid lower-precision
representation~\citep{anyprecision, matquant, matgptq, anybcq, rrq}.  The precision is chosen when the weights are loaded, and
nothing has to be requantized to change it.  Signal coding studied the
same property long before LLMs existed, under the name embedded
quantization~\citep{equitzcover, sullivan2004}.  Embedded quantization was used mostly in image coding, and
JPEG2000~\citep{pearlman2008wavelet} is the classical instance.  An LLM weight path differs from image coding in its source statistics,
group granularity, metadata budget and decoder.  In each of these respects the prefix property carries a cost.  The recent LLM formats, which are in effect variants
of classical embedded quantization, pay these costs.

Another established technique for low-bit accuracy holds a few
extreme weights out of the grid~\citep{squeezellm, spqr}.  In single-precision PTQ the held-out weights are chosen for one precision.  In a multi-precision format, however, a weight set aside changes what every precision reads.  Which weights to hold out therefore becomes part of designing the base grid.

We start from the embedded-quantization literature rather than from
the recent formats, because it already lists the choices involved in designing a grid that
is read at several precisions.  The constraints of the LLM weight path settle which of those choices
are open.  We make the open choices under those constraints and obtain
\stagq, a multi-precision weight format with two parts.  A
sparse side record holds the extreme weights out of the grid.  The
record is filled in two stages, by a pre-grid guard before the grid is
fitted and by a post-grid guard after it is fixed.  The main stream is a
group-wise affine 2-bit base refined by signed binary planes on a
dyadic step schedule (Figure~\ref{fig:ladder-example}).  The base and plane fields read at any supported precision concatenate
into one ordinary integer. One affine map then reconstructs the main-stream weight by scaling that
integer by a step and adding an offset.  Both are derived from one pair stored per group, and no per-weight lookup is needed.  Because the fields
are stored in plane order, loading a lower precision fetches fewer bytes.

On Llama-3.1-8B we evaluate two configurations, \stagxxs and \stagxs,
against Any-Precision LLM~\citep{anyprecision},
AnyBCQ~\citep{anybcq} and MatGPTQ~\citep{matgptq}. On Phi-4 and
OLMo-2-7B we compare \stagxxs.  At two bits it leads the strongest multi-precision baseline on MMLU on all three
models.  At three bits it leads on Llama-3.1-8B, leads on Phi-4 while
spending more bits, and ties on OLMo-2-7B.  Our kernel is faster than the Any-Precision LLM and AnyBCQ kernels in most shape--precision cases of
a batch-one matrix--vector product on an NVIDIA A100 GPU, timed on
synthetic weights.

\section{Background}
\label{sec:related-work}
\label{sec:background}

\subsection{Multi-Precision Quantization}
\label{sec:related-multiprecision}

\label{sec:related-embedded}
A multi-precision weight format is an \emph{embedded}, or
multiresolution, scalar quantizer.  Reading a longer prefix refines
the same scalar.  Under the same rate definition, distortion measure, calibration
information and decoder class, every prefix-readable code is also a
feasible fixed-precision code, but not conversely.  The distortions are therefore ordered, and a prefix-readable format cannot beat the best
fixed-precision quantizer at matched rate.

Scalar quantization~\citep{max1960, lloyd1982, gishpierce} and its embedded
form~\citep{equitzcover, brunkfarvardin, sullivan2004, pearlman2008wavelet}
have been studied extensively.  In the rate--distortion limit, successive refinability is the case where that ordering holds with equality.  \citet{equitzcover} show that the equality holds only for particular sources and distortion measures.

The recent LLM formats are variants of this classical object and fall
into three families by mechanism.  \emph{Slice nesting} takes
a lower precision from the leading digits of a deeper code.
MatQuant~\citep{matquant} achieves this nesting through training, and
\mbox{MatGPTQ}~\citep{matgptq} through post-training quantization.  Both anchor at the highest precision, and a shallow read inherits its
constants.  Such a read covers a fraction of the highest-precision
range, and its clipped end cell folds into a wider one.  SliceMoE~\citep{slicemoe} truncates an affine code and
its integer zero-point by the same bit shift.  \emph{Plane and residual
nesting} adds a separately scaled term per precision.
AnyBCQ~\citep{anybcq} reuses previously assigned binary codes,
RRQ~\citep{rrq} quantizes the residual left by the preceding stages, and
D2MoE~\citep{d2moe} pairs an affine base with binary refinement.  Their
metadata grows with the number of precisions exposed.
Single-precision BPDQ~\citep{bpdq} fits an independent scale per bit plane.  Its feasible level vectors fill a two-dimensional family, whereas one
scale spans only a ray.  A format that decodes every precision from one
stored pair is confined to that ray.  LUT-GEMM~\citep{lutgemm} already expresses uniform quantization on that
ray, as binary planes with power-of-two scales, but feeds the planes to
a lookup-table matrix kernel.  \emph{Learned
lookup reconstruction} admits irregular level sets.  Any-Precision
LLM~\citep{anyprecision} quantizes to its lowest precision first and
reaches each higher precision by splitting the clusters of that first
fit, storing one refitted lookup table per output channel per precision.
The refitted tables require an indexed reconstruction and metadata that grows with the precisions
served.

\stagq, proposed in this work, decodes every precision from one stored pair and reads its planes
by prefix, reconstructing each shallower read at the midpoint of the
cells it merges.  It fits that pair jointly across every precision the
stream serves.

\subsection{Designing the Grid}
\label{sec:constraints}
\label{sec:grid}

The classical direct-sum construction of residual
quantization~\citep{barnes1996} reconstructs a value from a
prefix of one stream, adding one term per plane to a base term:
\begin{equation}
  \widehat W^{(M)}
  = c^{(M)} + \sum_{k=1}^{M} \alpha_k\,\varphi_k(d_k),
  \qquad 1 \le M \le N .
  \label{eq:direct-sum}
\end{equation}

The stream supports precisions up to \(N\).  A read at precision \(M\)
reconstructs a weight \(W\) as \(\widehat W^{(M)}\) from \(M\) of the
\(N\) planes.  Quantization is per group, with \(g\) weights sharing one set of stored
constants.  The base term \(c^{(M)}\) sets the origin at the precision being read.
The step \(\alpha_k\) scales the term that plane \(k\) adds. The digit
map \(\varphi_k\) gives the value of the digit \(d_k\) at that plane.  Designing
a grid means choosing all three.  The weights constrain the choice
through their statistics, while the serving system constrains it
through the metadata it can store and the decoder it can run.  Bit-plane image coders such as JPEG2000~\citep{pearlman2008wavelet} use
this construction (Appendix~\ref{app:jpeg}).

Let \(u\) be the integer represented by the fields read at precision
\(M\).  A
group-wise affine grid at precision \(M\) has step \(s_M\) and origin
\(lo_M\), the value that code \(u=0\) takes under midpoint
reconstruction.  An interior code \(u\) corresponds to the half-open
cell \(C_u\) of width \(s_M\) centered on \(lo_M+u\,s_M\), whose lower edge
we write \(a_u\).  For an interior cell,
\begin{equation}
  r_u=\mathbb{E}[W\mid W\in C_u],
  \qquad
  \delta_u=\frac{r_u-a_u}{s_M}
  \label{eq:conditional-centroid}
\end{equation}
are the conditional mean of that cell and its position
within the cell, with \(\delta_u=\tfrac12\) at the midpoint.
Equation~\eqref{eq:direct-sum}, with \(c^{(M)}\) allowed to vary with
the code read, takes the form
\begin{equation}
  \widehat W^{(M)}
  =lo_M+(u+\epsilon_u)s_M ,
  \qquad
  \epsilon_u=\delta_u-\tfrac12 .
  \label{eq:affine-shape-centered}
\end{equation}
The same three choices reappear here in affine coordinates. The base
term \(c^{(M)}\) gives \(lo_M\) together with the displacement
\(\epsilon_u\), the steps \(\{\alpha_k\}\) give the spacing \(s_M\), and
the digit maps \(\{\varphi_k\}\) give the integer \(u\). The next
section fixes all three for \stagq in this order.

\section{The \stagq Framework}
\label{sec:format}

The encoder splits the weights into two parts.  A sparse side record \(P\) carries the entries that the two guards hold
out of the quantizer, one before the grid is fitted and one after.
The pre-grid guard records the value it removes, and the post-grid guard
records the residual the fixed grid leaves.  The rate charges
each held-out entry for its position.  The main stream, the multi-precision part, carries everything else.  The complete reconstruction is
\begin{equation}
  \widehat W^{(M)} = (1-m)\odot\widehat W_{\mathrm{main}}^{(M)} + P ,
  \label{eq:constraints-complete}
\end{equation}
where \(m\) marks the held-out entries.  At a pre-grid entry, \(P\) is the
recorded value.  At a post-grid entry it is the two-bit reconstruction
plus the recorded residual.  Neither depends on \(M\), and one \(P\)
therefore serves every precision.
Both the grid design and the prefix property belong to the
main stream, which carries a two-bit base followed by one-bit
refinement planes.  The evaluated instance sets \(N=6\), exposing five read precisions,
\(p2\) through \(p6\). Each added plane raises the payload by one bit
per weight.  A larger \(N\) only appends further planes and
adds no metadata.

\subsection{Reconstruction Points from Measured Weights}\label{sec:source}

The first choice to fix is the base term \(c^{(M)}\) of
Equation~\eqref{eq:direct-sum}, which sets the reconstruction point
inside the cell selected by the fields read at precision \(M\).  For fixed boundaries and squared error, the best position is the conditional centroid \(r_u\) of
Equation~\eqref{eq:conditional-centroid}.  A closed form for \(r_u\) requires a density.  The \(\delta\approx0.38\)
of JPEG2000~\citep{pearlman2008wavelet}, for example, assumes a
Laplacian one.  Grouped LLM weights
have no established density of that kind, and we therefore measure
\(r_u\) on the checkpoint weights.  The measurement partitions each group with the conventional lattice
of four levels, the outer two on the group minimum and maximum
(Appendix~\ref{sec:appendix-models}).

\begin{figure}[t]
  \centering
  \includegraphics[width=\linewidth]{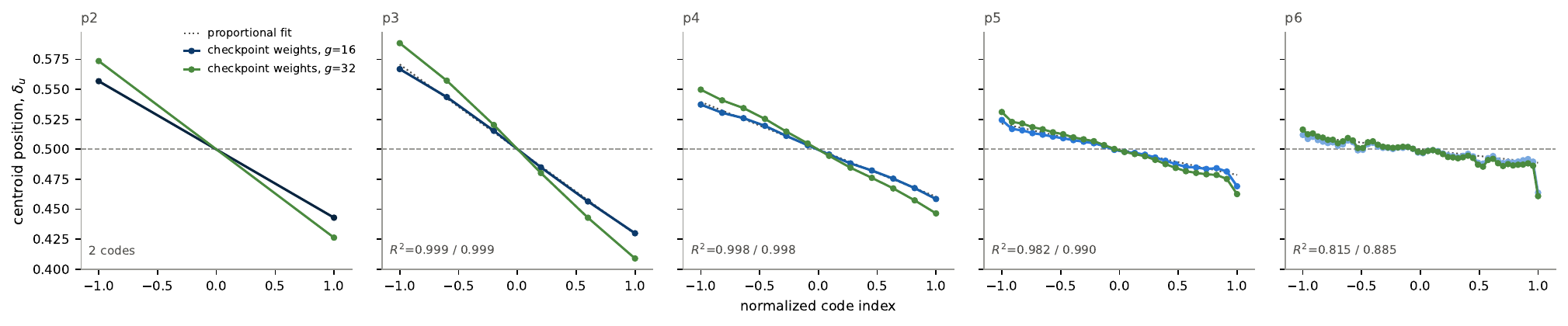}
  \caption{Measured position \(\delta_u\) of the conditional centroid
  inside each code cell, at five precisions on a common vertical
  scale.  Blue: Llama-3.1-8B at \(g=16\); green: \(g=32\).  Further
  checkpoints are in Appendix~\ref{sec:appendix-models}.}
  \label{fig:delta-shape}
\end{figure}

Figure~\ref{fig:delta-shape} plots the measured \(\delta_u\).  The dashed line marks the midpoint, where a flat within-cell density
puts the centroid.  A uniform control recovers the midpoint at every precision and
at both group sizes \stagq uses, \(g=16\) and
\(g=32\).  Because only two cells remain at \(p2\), the shape is read from \(p3\) on.
Through \(p5\), the centroids fall on a straight line that crosses
\(\tfrac12\) at the group center.  Each centroid is displaced toward the center in proportion to its
distance from the center.  A displacement of that form needs no per-code correction, since at any
one precision a rescaling of the grid absorbs it.

The pair \((lo,s)\) of each group is quantized with two sixteen-entry
codebooks fitted per matrix, one in \(\log_2 s\) and one in \(lo/s\).
Together they span \(256\) combinations, and each group then takes the
combination that minimizes the loss across every precision the stream
serves.  For weights \(W_c\) reconstructed at precision \(M\) as
\(\widehat W_c^{(M)}\), and with every precision weighted equally,
\begin{equation}
  D(lo,s)=\frac{1}{N-1}\sum_{M=2}^{N}\ \sum_{c}
          E_c\bigl(W_c-\widehat W_c^{(M)}\bigr)^{2}
          \;+\;\lambda\,\Phi(lo,s),
  \label{eq:ladder-objective}
\end{equation}
where \(E_c\) is the calibration mean square of input channel \(c\)
(column \(c\) of \(W\)), and \(\Phi\) is a floor term,
\begin{equation}
  \Phi(lo,s)=\sum_{c} E_c
  \max\bigl(|W_c-\widehat W_c^{(2)}|-\rho,\;0\bigr)^{2},
  \label{eq:floor-term}
\end{equation}
in which \(\rho=s\sum_{k=3}^{N}2^{-(k-1)}\) is the furthest the
refinement planes can move the base reconstruction.  We set \(\lambda=4\).  The sum runs over the whole
group, with any entry the pre-grid guard holds out clipped to the next
value inward.    The floor term \(\Phi\) charges only the part of the base
error that no plane can reach.  Because the coarse two-bit read dominates the first term, that term alone would underweight an error whose damage shows at the fine reads.

Across the seven projections of four sampled layers of Llama-3.1-8B, the
fitted grids clip a few extreme weights in exchange for finer levels across
the body of each group.  In these fitted grids, no level sits on a group
extreme.

\subsection{Step Schedules under a Metadata Budget}\label{sec:side-info}

The second choice to fix is the refinement step \(\alpha_k\) of
Equation~\eqref{eq:direct-sum}, for each plane \(k\geq3\).  To avoid fitting and storing one scale per
refinement stage, \stagq fits only the base step \(s\) and derives every
refinement step from it.

Suppose a refinement bit splits an interval cell into two interval
children, and the density inside the parent cell is uniform at the
scale of the split.  The minimum-squared-error design is then the
midpoint split.
Applied at every plane, this gives the dyadic schedule of refinement half-steps
\begin{equation}
  t_k=\alpha_k=\frac{s}{2^{k-1}},
  \qquad k=3,\ldots,N .
  \label{eq:dyadic-schedule}
\end{equation}
At precision \(M\) the cell spacing is \(s_M=s/2^{M-2}\), twice the
additive half-step \(t_M\).  Appendix~\ref{app:dyadic} states when this schedule is optimal.  On Llama-3.1-8B, a global rescale \(\gamma\) of its steps lowers the median output error by at most \(0.6\%\).  The released format keeps \(\gamma\) at \(1\), which leaves the schedule of Equation~\eqref{eq:dyadic-schedule} unchanged.

\subsection{Digit Maps and Layout for the Kernel}\label{sec:kernel}

The third choice is the digit maps \(\varphi_k\), which translate stored
bits into numbers, and we fix them together with the physical layout.
Both follow from the decode interface of an LLM, which requires fixed-width
addressability, affine decoding with no per-weight lookup, and the
prefix property.
Fixed-width addressability restricts each field to a fixed alphabet
\(\mathcal A_k=\{0,\ldots,2^{m_k}-1\}\), with \(m_k\) a format constant.  A field's location is then an affine
function of the weight index, and its unpack width is known when the
kernel is built.

Affine decoding with no per-weight lookup restricts each digit map to
an affine function.  Our instance takes a two-bit base, \(q\in\{0,1,2,3\}\), with
the identity as its digit map, followed by one-bit refinement digits
\(b_k\in\{0,1\}\), each encoding the sign of the current residual.
Therefore, with the half-steps \(t_k\) of
Equation~\eqref{eq:dyadic-schedule}, each digit map is
\(\varphi_k(b)=2b-1\), and plane \(k\) then adds \(2t_kb_k-t_k\).  The constants these planes accumulate depend only on the
precision read and the base step, and thus none of them is stored.  A zero residual takes \(\operatorname{sign}(0):=+1\), and the
reconstruction still moves by one half-step.  Because the fitted grids
of Section~\ref{sec:source} keep the levels off the group extremes, a
ternary alphabet is never needed.  The two bits of \(q\) followed by
\(b_3,\ldots,b_M\) form the composite integer
\begin{equation}
  u_M=q\,2^{M-2}+\sum_{k=3}^{M}b_k\,2^{M-k},
  \label{eq:composite-integer}
\end{equation}
which is the abstract \(u\) of Equation~\eqref{eq:affine-shape-centered}.
Main-stream reconstruction is therefore a scale-and-offset of one
integer.

To keep the prefix property in the bytes, \stagq stores the base first
and then the refinement planes in order of \(k\).  A shallower load
therefore skips the deeper planes.  Our kernel instead reads contiguous
\(M\)-bit digits, repacked once at load for one
precision (Appendix~\ref{app:kernel-impl}).

\subsection{The Resulting \stagq Main Stream}\label{sec:stagq}

With all three choices fixed, substituting the two-bit base, the signed one-bit refinements, and the
unscaled dyadic schedule into the additive reconstruction of
Equation~\eqref{eq:direct-sum} gives
\begin{equation}
  \widehat W_{\mathrm{main}}^{(M)}
  =lo+s q+\sum_{k=3}^{M}(2b_k-1)\frac{s}{2^{k-1}},
  \qquad M\in\{2,\ldots,N\},
  \label{eq:main-stream-additive}
\end{equation}
with the sum empty at \(M=2\).  All data-dependent terms are integer
multiples of the half-step \(t_M=s/2^{M-1}\) of the deepest plane read,
and they combine into the composite integer \(u_M\) of
Equation~\eqref{eq:composite-integer}.  With \(s_M=s/2^{M-2}\) and
\(lo_M=lo-s(2^{M-2}-1)/2^{M-1}\), the reconstruction collapses to
\begin{equation}
  \widehat W_{\mathrm{main}}^{(M)}
  =lo_M+s_Mu_M .
  \label{eq:main-affine}
\end{equation}
Every precision therefore decodes as one affine map of one ordinary
integer, with \((lo_M,s_M)\) derived in closed form from the single
stored pair \((lo,s)\).  This map covers the main stream only.  The side record \(P\) of Equation~\eqref{eq:constraints-complete} is
applied outside this map, and a complete runtime must fuse \(P\) or
report its traffic separately.

\subsection{Holding Out the Extremes}\label{sec:holdout}

How to treat the extreme weights must be settled before the pair of
Equation~\eqref{eq:ladder-objective} is fitted, because one embedded
grid serves every precision.
Consider a group with one weight far larger than the rest.  Widening
the pair \((lo,s)\) to cover that weight spreads the four base levels
apart, so every precision decodes the group more coarsely.  Instead,
fitting the pair to the rest leaves the extreme weight past the
outermost of those four levels, where the base clips it.  Because the planes reach at most \(\rho\) past that level
(Equation~\eqref{eq:floor-term}), no precision can recover a weight
lying further out.  For example, on Phi-4's \(80\) attention
matrices at \(g{=}32\), the entries \stagxxs selects remove \(26.9\%\) of the
two-bit error when held out before the grid is fitted, but only the
\(6.2\%\) they carry when held out after it (Appendix~\ref{app:encoder}).

The pre-grid guard takes at most one entry per group, under one of two criteria that the configuration assigns to each projection.  The ratio criterion holds out a group's largest-magnitude entry when it exceeds the second largest by more than a factor of \(T\), four on Llama-3.1-8B.  The gain criterion weighs the two-bit error an entry's removal saves against the storage the entry occupies.
The post-grid guard runs after the
grid is fixed and takes a second set, scored on the error that
grid leaves behind.

To score a candidate at position \(j\), the gain criterion clips it to
the next value inward and refits the matrix codebooks and pairs.  The
gain \(\Delta_j\) is the fall this causes in the group's two-bit error
\(D_2\), and the candidate is held out when its gain per bit of storage
exceeds a threshold \(\tau\):
\begin{equation}
  \Delta_j \;=\; D_2(V)-D_2\bigl(V^{j\to}\bigr) , \qquad
  \Delta_j \,/\, b \;>\; \tau .
  \label{eq:gain-criterion}
\end{equation}
Here \(V\) holds the group's values, \(V^{j\to}\) the same values with
the candidate clipped, and \(b\) the bits the entry occupies.  One \(\tau\) per stream serves both guard stages.
Because \(\Delta_j\) includes what the refit recovers and depends on the other groups' candidates, it is a heuristic score.

\subsection{The Encoder}
\label{sec:evaluation-implementation}

\begin{figure}[t]
  \centering
  \includegraphics[width=\linewidth]{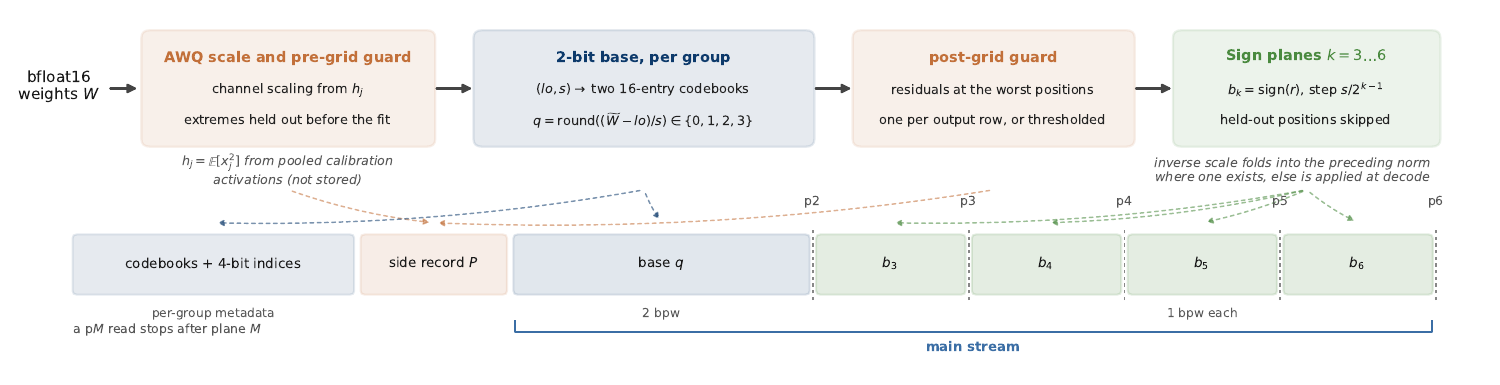}
  \caption{The \stagq encoder and the fields its rate charges.  Dashed
lines in each stage's color map the stage to the field it fills.  A
p\(M\) read stops after plane \(M\).}
  \label{fig:encoder-pipeline}
\end{figure}

The encoder first rescales each input channel with an AWQ-style
activation-aware scale~\citep{awq}.  A calibration pass measures the mean square of each
channel's activations, and channels with larger activations are scaled
up so that the grid resolves their weights more finely.  The quantizer
then operates on the scaled matrix.  Columns are partitioned into groups of \(g\), and each
group's pair is fitted as Section~\ref{sec:source} describes.  The two
guard stages of Section~\ref{sec:holdout} hold weights out of the
quantizer, following standard practice~\citep{squeezellm, spqr}.  Each
refinement plane \(k=3,\ldots,6\) stores the sign of the residual left
by the preceding prefix and skips every position the side record
already carries.  No deeper prefix can therefore overwrite a held-out
entry.  Figure~\ref{fig:encoder-pipeline} maps every stage to the field
it fills.

Every read carries the whole side record.  On Llama-3.1-8B, for example, \stagxs spends \(0.13\) bits per weight on the entries its gain criterion holds out, the same at every read.  A deeper read cannot drop them, because the pair was fitted with the extremes clipped, and the main
stream there encodes only the clipped value.  A quantizer built for six bits alone would extend its grid over these
weights and store no record at all.  This overhead is the cost of serving every precision from one checkpoint.

\section{Experimental Results}
\label{sec:evaluation}

\subsection{Experimental Settings}
\label{sec:evaluation-protocol}

We report two encoder configurations, \stagxs and the lower-rate
\stagxxs.  At two bits \stagxs runs at
\(2.51\) to \(2.64\) bits per weight (\bpw) and \stagxxs at \(2.34\)
to \(2.35~\bpw\), within \(0.04~\bpw\) of the strongest two-bit baseline on
all three models.  Each refinement plane adds exactly \(1~\bpw\).  Appendix~\ref{app:configurations}
gives the group sizes, codebook-fitting constants, scaling exponents
and the pre-grid guard settings of both configurations.

We evaluate on three dense models, Llama-3.1-8B~\citep{llama3},
Phi-4~\citep{phi4} and OLMo-2-7B~\citep{olmo2}.  Both configurations are reported on Llama-3.1-8B.  Every linear layer of the transformer blocks is
quantized.  The embedding and the language-model head stay in bfloat16,
as in the baselines' releases.  We report 5-shot
MMLU~\citep{mmlu} and WikiText-2~\citep{wikitext} perplexity, and on
Llama-3.1-8B also five zero-shot commonsense-reasoning (CSR) tasks with
their average (the CSR average).  We ran each entry ourselves under one
harness, in \texttt{bfloat16}.  Every perplexity we report is
token-normalized over non-overlapping \(2{,}048\)-token blocks, as in
the baselines' releases.  These token-level perplexities are
therefore not comparable to a word-level perplexity.  The
baselines are MatGPTQ,
Any-Precision LLM and AnyBCQ, each read from one multi-precision checkpoint by
a prefix cut.  Rate is the logical \bpw of the payload, the side record and every
side-information field a format declares, computed from the declared
widths.  The \stagq rate also charges
two mechanisms no baseline here uses, the finer groups and the two guard
stages, whose settings are chosen per model and configuration, with the
hold-out thresholds also set per block.  On
OLMo-2-7B we chose the thresholds by rate alone.  On
Llama-3.1-8B and Phi-4 we also compared candidate settings on WikiText-2
perplexity and MMLU, and on Llama-3.1-8B also on the CSR average.  At two bits on
Llama-3.1-8B our kernel, loaded with real weights, reads \(2.47~\bpw\)
against the \(2.35\) charged.  Most of the gap comes from sizing the
per-row guard slots of each feed-forward matrix to its fullest row
(Appendix~\ref{app:limitations}).

\begin{table}[t]
\centering
\caption{Accuracy on Llama-3.1-8B, in AnyBCQ's column order (HS is HellaSwag and WG WinoGrande).  Bit is the read precision \(M\) and Rate the logical \(\bpw\).
Bold marks the best quantized entry per
column within a precision.}
\label{tab:dense-csr}
\footnotesize
\setlength{\tabcolsep}{4.0pt}
\begin{tabular}{llrrrrrrrr}
\toprule
Method & Bit & Rate & MMLU & ARC-C & ARC-E & HS & PIQA & WG & CSR Avg \\
\midrule
bfloat16 & 16 & 16.00 & 65.23 & 53.67 & 81.06 & 78.91 & 81.28 & 73.72 & 73.73 \\
\midrule
MatGPTQ & 2 & 2.25 & 24.65 & 24.74 & 24.45 & 26.29 & 51.41 & 49.17 & 35.21 \\
Any-Precision & 2 & 2.01 & 23.82 & 24.57 & 34.09 & 27.90 & 57.24 & 50.36 & 38.83 \\
AnyBCQ & 2 & 2.38 & 35.33 & 37.29 & 62.67 & 62.76 & 73.94 & 57.93 & 58.92 \\
\stagxxs & 2 & 2.35 & 38.38 & 38.05 & 60.94 & 63.63 & 73.23 & 65.27 & 60.23 \\
\stagxs & 2 & 2.64 & \textbf{48.67} & \textbf{41.98} & \textbf{70.12} & \textbf{70.05} & \textbf{75.52} & \textbf{67.32} & \textbf{65.00} \\
\midrule
MatGPTQ & 3 & 3.25 & 44.84 & 42.41 & 67.97 & 68.39 & 74.05 & 67.32 & 64.03 \\
Any-Precision & 3 & 3.03 & 55.69 & 47.27 & 74.58 & 73.36 & 78.67 & 67.72 & 68.32 \\
AnyBCQ & 3 & 3.50 & 58.13 & 46.93 & 75.88 & 74.37 & 79.16 & 69.46 & 69.16 \\
\stagxxs & 3 & 3.35 & 61.89 & 49.32 & 76.60 & 76.57 & 79.65 & \textbf{72.45} & 70.92 \\
\stagxs & 3 & 3.64 & \textbf{62.55} & \textbf{51.02} & \textbf{77.48} & \textbf{77.71} & \textbf{80.41} & 71.43 & \textbf{71.61} \\
\midrule
MatGPTQ & 4 & 4.25 & 59.29 & 48.98 & 76.43 & 75.57 & 78.94 & 71.11 & 70.21 \\
Any-Precision & 4 & 4.05 & 63.55 & \textbf{52.73} & \textbf{80.26} & 78.21 & 80.79 & 72.06 & 72.81 \\
AnyBCQ & 4 & 4.63 & 63.17 & 51.71 & 78.91 & 77.28 & 80.47 & 73.09 & 72.29 \\
\stagxxs & 4 & 4.35 & 64.11 & 52.56 & 79.84 & 78.63 & 80.47 & 73.24 & 72.95 \\
\stagxs & 4 & 4.64 & \textbf{64.44} & 52.30 & 80.18 & \textbf{79.11} & \textbf{80.85} & \textbf{73.88} & \textbf{73.26} \\
\midrule
Any-Precision & 5 & 5.10 & 64.94 & \textbf{53.07} & \textbf{81.14} & 78.74 & 81.18 & 73.01 & 73.43 \\
\stagxxs & 5 & 5.35 & 64.73 & 52.56 & 80.56 & 78.94 & \textbf{81.45} & 72.93 & 73.29 \\
\stagxs & 5 & 5.64 & \textbf{65.07} & 52.99 & 80.98 & \textbf{79.33} & 80.74 & \textbf{73.40} & \textbf{73.49} \\
\midrule
Any-Precision & 6 & 6.20 & \textbf{65.20} & 53.41 & 81.19 & 79.03 & \textbf{81.18} & 73.56 & 73.67 \\
\stagxxs & 6 & 6.35 & 64.83 & 52.99 & 80.68 & 78.95 & 81.07 & 73.40 & 73.42 \\
\stagxs & 6 & 6.64 & 65.07 & \textbf{53.75} & \textbf{81.23} & \textbf{79.40} & 81.01 & \textbf{73.88} & \textbf{73.85} \\
\bottomrule
\end{tabular}
\end{table}

\subsection{Accuracy Evaluation}
\label{sec:accuracy}

At two and three bits, \stagxxs scores higher on MMLU than the
strongest baseline at that precision on all three models
(Tables~\ref{tab:dense-csr} and~\ref{tab:phi4-acc}).  However, its
\(0.9\)-point lead on OLMo-2-7B at three bits falls inside our one-point tie band on MMLU.  \stagxxs is also cheaper than that baseline in five of
these six model--precision cases.  The exception is Phi-4 at three bits,
where it costs about a third of a bit more than Any-Precision LLM and
leads it by \(1.2\) MMLU points, an upper bound on the equal-rate
advantage.  On the Llama-3.1-8B CSR average at two and
three bits, \stagxxs leads the strongest baseline by \(1.3\) and \(1.8\)
points, outside our tie band of \(0.6\) points on that average.  On Llama-3.1-8B at two bits, \stagxs costs a
quarter of a bit more than AnyBCQ and leads it by \(13.3\) MMLU points,
again an upper bound on the equal-rate advantage.  These comparisons
are between complete encoder configurations.  They do not separate the
format from the finer groups and the guards, which no baseline here
uses.  The groups of \(32\) stay cheap because each group stores one
eight-bit pair that every precision reads.  This pair costs
\(0.25~\bpw\), less than the \(0.375\) to \(0.625~\bpw\) that AnyBCQ's
per-precision records cost at two to four bits with groups of \(128\).

From four bits up the strongest baseline is Any-Precision LLM, which is
\(0.15\) to \(0.30~\bpw\) cheaper than \stagxxs.  At four bits \stagxxs
leads it by under one MMLU point.  At five and six bits on Llama-3.1-8B
it trails by up to \(0.4\) points on MMLU and on the CSR average.  We count all of these as ties.  \stagxxs has lower
perplexity on the C4~\citep{c4} validation shard than every baseline at two to four bits on Llama-3.1-8B and
OLMo-2-7B.  In two of these cases the best
baseline on C4 is cheaper.  At five and six bits on Llama-3.1-8B,
Any-Precision LLM has a lower rate and a lower C4 perplexity.

\FloatBarrier

\begin{table}[t]
\centering
\caption{MatGPTQ (MatQ), Any-Precision LLM (AP), AnyBCQ (AB) and
\stagxxs (XXS).  Tok/s is the mean of six timings in one run.
OLMo-2-7B timings omit every per-channel multiply, favoring us.}
\label{tab:phi4-acc}
\scriptsize
\setlength{\tabcolsep}{2.6pt}
\begin{tabular}{lrrrrr@{\hskip 8pt}rrrrr@{\hskip 8pt}rrrr}
\toprule
& \multicolumn{5}{c}{Wiki \(\downarrow\)} & \multicolumn{5}{c}{MMLU \(\uparrow\)}
& \multicolumn{4}{c}{tok/s \(\uparrow\)} \\
\cmidrule(lr){2-6}\cmidrule(lr){7-11}\cmidrule(lr){12-15}
Bit & bf16 & MatQ & AP & AB & XXS & bf16 & MatQ & AP & AB & XXS
    & fp16 & AP & AB & XXS \\
\midrule
\multicolumn{15}{l}{\emph{Llama-3.1-8B}} \\
2 & 6.24 & \(>\)10\(^{6}\) & 4674.38 & 19.03 & \textbf{12.77}
  & 65.23 & 24.65 & 23.82 & 35.33 & \textbf{38.38}
  & 46.16 & 73.91 & \textbf{74.15} & 73.68 \\
3 & 6.24 & 11.07 & 8.58 & 8.09 & \textbf{7.24}
  & 65.23 & 44.84 & 55.69 & 58.13 & \textbf{61.89}
  & 46.16 & 70.10 & 70.16 & \textbf{70.88} \\
4 & 6.24 & 7.48 & 6.70 & 6.84 & \textbf{6.57}
  & 65.23 & 59.29 & 63.55 & 63.17 & \textbf{64.11}
  & 46.16 & 66.32 & 66.36 & \textbf{68.31} \\
\midrule
\multicolumn{15}{l}{\emph{Phi-4}} \\
2 & 6.46 & \(9.4{\times}10^{5}\) & 17.71 & 9.69 & \textbf{8.84}
  & 80.34 & 24.65 & 46.38 & 66.66 & \textbf{70.68}
  & 29.00 & 52.98 & \textbf{54.40} & 53.86 \\
3 & 6.46 & 7.23 & 7.17 & 7.63 & \textbf{6.80}
  & 80.34 & 75.76 & 77.26 & 76.76 & \textbf{78.41}
  & 29.00 & 49.74 & \textbf{50.65} & 50.55 \\
4 & 6.46 & 6.71 & 6.63 & 7.32 & \textbf{6.54}
  & 80.34 & 79.29 & 79.50 & 79.01 & \textbf{79.92}
  & 29.00 & 46.22 & 47.20 & \textbf{48.81} \\
\midrule
\multicolumn{15}{l}{\emph{OLMo-2-7B}} \\
2 & 6.10 & \(4.8{\times}10^{5}\) & \(5.5{\times}10^{4}\) & 11.74 & \textbf{8.98}
  & 62.78 & 22.95 & 25.51 & 46.03 & \textbf{53.01}
  & 50.50 & 82.12 & 83.88 & \textbf{85.50} \\
3 & 6.10 & 8.11 & 7.35 & 7.28 & \textbf{6.68}
  & 62.78 & 57.68 & 57.53 & 59.15 & \textbf{60.08}
  & 50.50 & 78.16 & 79.46 & \textbf{80.49} \\
4 & 6.10 & 6.66 & 6.38 & 6.48 & \textbf{6.24}
  & 62.78 & 61.47 & 61.91 & 61.44 & \textbf{62.68}
  & 50.50 & 73.64 & 75.13 & \textbf{78.44} \\
\bottomrule
\end{tabular}
\end{table}

\begin{table}[t]
\centering
\caption{Guard ablation, Llama-3.1-8B, \(g{=}32\).  The side-record column is the rate the guards add.
Every row stores the payload bits and \(0.25~\bpw\) of
per-group pairs.}
\label{tab:ablation}
\scriptsize
\setlength{\tabcolsep}{3.2pt}
\begin{tabular}{lr@{\hskip 10pt}rrrrr@{\hskip 10pt}rrrrr}
\toprule
& side record & \multicolumn{5}{c}{Wiki \(\downarrow\)} & \multicolumn{5}{c}{MMLU \(\uparrow\)} \\
\cmidrule(lr){3-7}\cmidrule(lr){8-12}
Bit & (\bpw) & 2 & 3 & 4 & 5 & 6 & 2 & 3 & 4 & 5 & 6 \\
\midrule
neither guard  & 0      & 28.47 & 7.60 & 6.66 & 6.43 & 6.38 & 26.90 & 58.03 & 62.62 & 64.29 & 64.74 \\
post-grid only & 0.0056 & 23.18 & 7.47 & \textbf{6.58} & \textbf{6.36} & \textbf{6.31} & \textbf{28.51} & 58.63 & 63.11 & 64.47 & \textbf{64.88} \\
pre-grid only  & 0.0010 & 24.77 & 7.57 & 6.69 & 6.48 & 6.43 & 25.00 & 59.74 & 63.84 & 64.40 & 64.78 \\
both           & 0.0065 & \textbf{20.86} & \textbf{7.44} & 6.60 & 6.41 & 6.36 & 25.09 & \textbf{60.30} & \textbf{64.26} & \textbf{64.64} & 64.77 \\
\bottomrule
\end{tabular}
\end{table}

The guards matter most at low bits (Table~\ref{tab:ablation}).
On a Llama-3.1-8B stream held at \(g{=}32\) throughout, both guards
together lower the two-bit perplexity from \(28.47\) to \(20.86\), still
behind AnyBCQ's \(19.03\) but at a lower rate.  Two-bit MMLU is near chance in every row.
The guards add \(2.3\) and \(1.6\) MMLU points at three and four bits,
and only \(0.35\) and \(0.03\) at five and six.  Every read at those precisions
still carries their side record.

\begin{wraptable}{r}{0.44\linewidth}
\vspace{-\intextsep}
\centering
\caption{Single-precision control on Llama-3.1-8B.
\(\Delta\)MMLU is the dedicated fit minus \stagxxs.}
\label{tab:single-precision}
\small
\setlength{\tabcolsep}{4pt}
\begin{tabular}{lrrr}
\toprule
& \multicolumn{2}{c}{Wiki \(\downarrow\)} & \\
\cmidrule(lr){2-3}
Bit & \stagxxs & dedicated & \(\Delta\)MMLU \\
\midrule
2 & 12.77 & 11.99 & \(+3.62\) \\
3 & 7.24  & 7.18  & \(-0.03\) \\
4 & 6.57  & 6.44  & \(+0.72\) \\
\bottomrule
\end{tabular}
\end{wraptable}

Table~\ref{tab:single-precision} measures the accuracy cost of the
prefix constraint only within our own encoder, on Llama-3.1-8B.  We refit
\stagxxs for one precision at a time and hold every other setting fixed.  At two bits the dedicated fit spends slightly more rate and scores
\(3.6\) MMLU points higher, an upper bound on the equal-rate cost.  At three and four bits the dedicated
fit is cheaper by under \(0.01~\bpw\).  The MMLU differences there,
\(-0.03\) and \(+0.72\) points, bound the cost from below.  Both lie
inside our tie band.

\subsection{Kernel Evaluation}
\label{sec:gpu-reference}

Table~\ref{tab:gemv-a100} reports the median speedup of the three
quantized kernels over cuBLAS in a batch-one matrix--vector product on
the A100, across the nine shapes of AnyBCQ's kernel table.  Each
baseline runs at the group size of its released kernel, and \stagq at
\(g{=}32\).  The \stagq kernel is faster than the Any-Precision LLM
kernel in \(24\) of \(27\) shape--precision cases and than the AnyBCQ
kernel in \(23\).  All seven comparisons it loses fall on the three attention
shapes, the only ones timed with per-group guard space.  Omitting that
space on the six feed-forward shapes favors \stagq.  These timings use synthetic weights.  On a real two-bit Llama-3.1-8B stream,
our decode path raises perplexity on the first four WikiText-2 blocks by \(0.0086\) against the same weights in
\texttt{fp16}, well inside a \(0.10\) tolerance.

\FloatBarrier
\begin{wraptable}{r}{0.46\linewidth}
\vspace{-\intextsep}
\centering
\caption{Median speedup over cuBLAS, \(t_{\mathrm{cuBLAS}}/t_{*}\), on
synthetic weights.  The \stagq kernel here applies no activation scale.
Its layout reserves guard space only on the three attention shapes.
Appendix~\ref{app:gemv} gives the per-shape latencies.}
\label{tab:gemv-a100}
\small
\begin{tabular}{lrrr}
\toprule
Bit & Any-Precision & AnyBCQ & \stagq \\
\midrule
2 & \(3.53\times\) & \(3.93\times\) & \(\mathbf{4.13\times}\) \\
3 & \(2.76\times\) & \(3.01\times\) & \(\mathbf{3.39\times}\) \\
4 & \(2.34\times\) & \(2.43\times\) & \(\mathbf{3.00\times}\) \\
\bottomrule
\end{tabular}
\end{wraptable}

In Table~\ref{tab:phi4-acc}, we evaluate end-to-end decode speed.  Our
path fuses format-independent operators that the baseline paths run
separately, as in their releases.  With these fusions and all decoders
timed in one run, our path is faster in fourteen of the eighteen
comparisons.  Timed at three bits without
the fusions or the per-channel multiply, it is slower than the fastest baseline on all three models, by
\(0.3\), \(0.6\) and \(2.9\%\).  All three quantized decoders are
far faster than the \texttt{fp16} decoder, ours by a factor of \(1.48\) to
\(1.86\).

\section{Discussion and Conclusion}\label{sec:discussion}

\stagq serves every precision from one prefix-readable main stream,
decoded by one affine map with no per-weight lookup. Two guards hold out
the few weights the grid serves worst in a sparse side record. On
Llama-3.1-8B, Phi-4 and OLMo-2-7B, \stagxxs leads the strongest
multi-precision baseline on MMLU at two bits, at a slightly lower
logical rate. At three
bits it leads on Llama-3.1-8B, leads on Phi-4 at a higher rate, and ties
on OLMo-2-7B. From four bits up it ties Any-Precision LLM on MMLU at a
higher rate. On synthetic weights, our matrix--vector kernel is faster than
both baseline kernels in most shape--precision cases. With format-independent fusions and on synthetic weights, our decode path
is faster in fourteen of eighteen end-to-end comparisons.  Unfused at three bits, it trails
the fastest baseline. 

Future work will extend the evaluation to larger models and to
mixture-of-experts models, and will ablate the parts of the encoder one
at a time.  On the systems side, with further
engineering, the kernel can be taken into production serving, where one
deployment switches among precisions at run time.

\subsubsection*{Acknowledgments}
We used generative AI tools to help develop the conceptual framework and to
help write the proof of Appendix~\ref{app:dyadic}.  We also used them to
write the code, to interpret results, to draft and edit the text and to
produce the figures.  We reviewed all AI-assisted work, checked every number
from our own runs against the underlying result files, verified every
derivation independently of the tools, and take responsibility for the
final content of this paper.

\bibliographystyle{iclr2027_conference}
\bibliography{refs}

\appendix
\raggedbottom
\section{Limitations}\label{app:limitations}

\paragraph{Rate.}
Every rate we report counts the bits that the format's fields declare.
On Llama-3.1-8B at two bits the kernel loaded with real weights reads
\(2.47\) bits per weight against the \(2.35\) charged.  Our loader
stores the guard entries of the feed-forward matrices in per-row slots,
and every row of a matrix carries as many slots as its fullest row.  A side record
indexed by output row would read \(2.39\) bits per weight.  One that
also marks each attention group with a presence bit would read \(2.354\),
within \(0.004\) of the charge.  Both need a kernel that decodes records
of variable length.  We leave that engineering change to
future work.

\paragraph{Scope of the evidence.}
All quality results come from three dense models, and none from a mixture-of-experts model.  Most
of the encoder analysis rests on the \(28\) matrices of four sampled
layers of Llama-3.1-8B.  The GEMV and decode timings are batch-one on one A100 with
synthetic weights.  Real weights cost our kernel slightly more shared-memory
conflict than synthetic weights do.  At two bits on Llama-3.1-8B, decoding with real weights reads about a fifth of the \texttt{fp16} weight bytes.  By bytes alone, decoding could therefore run \(4.7\) times as fast as \texttt{fp16}.  However, the decoder we time reads fewer bytes still and decodes only \(1.6\) times as fast.

\paragraph{Comparisons.}
Because no configuration lands on a baseline's rate, every comparison with a baseline is made across a rate difference.  Several baseline runs also depart from their authors' releases.  We supplied the model configurations for Any-Precision LLM on Phi-4 and OLMo-2-7B and for AnyBCQ on OLMo-2-7B.  Outside its released recipe, we ran MatGPTQ asymmetrically and added two bits to the precisions it optimizes jointly.  We score AnyBCQ's released Phi-4
checkpoint, and our own AnyBCQ runs on OLMo-2-7B, with each model's own
tokenizer rather than with the tokenizer files the release's code saves
beside the checkpoint.  Scored with those saved files instead, our
AnyBCQ MMLU on Phi-4 comes within \(0.05\) points of the value
\citet{anybcq} report, against up to \(0.32\) points with Phi-4's own
tokenizer.  Our AnyBCQ run on OLMo-2-7B uses the per-device training batch size
of two from the release's README, and a run at batch size one differs
from it by at most \(0.17\) MMLU points.  Our Any-Precision LLM runs also disagree with
the numbers \citet{anybcq} report for it at two bits.  We obtain
WikiText-2 perplexity \(4674\) against their \(1681\) on Llama-3.1-8B,
and \(17.71\) against their \(13.37\) on Phi-4.  At three and four bits
our Any-Precision LLM perplexities agree with theirs within \(0.03\).

\paragraph{Ablations and controls.}
We set the group sizes and guard criteria per model and configuration, and the hold-out
thresholds per model, configuration and block (Appendix~\ref{app:configurations}).  We report no
ablation of \(\lambda\), of the AWQ exponents or of how the pair \((lo,s)\)
is fitted.  Apart from the single-precision control, every fit we report
weights the precisions of Equation~\eqref{eq:ladder-objective} equally.  Among the streams reported here, the only one with nothing held out is
the neither-guard row of Table~\ref{tab:ablation}, on Llama-3.1-8B at \(g{=}32\).  In the
same table, adding the pre-grid guard on top of the post-grid guard
raises WikiText-2 perplexity at four to six bits by \(0.02\) to \(0.05\).  Outside the table, the same addition raises C4 perplexity at three to six bits by \(0.09\) to \(0.11\).  Because the both-guards row also
spends more rate than the post-grid-only row, these increases would be at
least as large at equal rate.  The single-precision control covers
Llama-3.1-8B only and measures the cost of the prefix constraint within
our own encoder.  It therefore does not bound that cost against the best
fixed-precision quantizer.

\section{Encoder Configurations}
\label{app:configurations}

Table~\ref{tab:configurations} lists the encoder settings of the two
reported configurations.  Group sizes and guard criteria are set per
model and configuration.  Hold-out thresholds are also set per block
(attention or feed-forward).  Phi-4 fuses
\(q/k/v\) into one projection and gate/up into another, and its
attention block therefore carries one group size.

\begin{table}[ht]
\centering
\caption{Encoder settings per model and configuration.  Each pre-grid
guard column names the matrices under that criterion and its threshold.
A dash marks a criterion that no matrix uses.  The two
criteria are exclusive per matrix.  On Phi-4 the fused gate/up projection carries only the
post-grid guard.}
\label{tab:configurations}
\scriptsize
\setlength{\tabcolsep}{3pt}
\begin{tabular}{llllll}
\toprule
 & & & & \multicolumn{2}{c}{Pre-grid guard} \\
\cmidrule(lr){5-6}
Model & Configuration & Feed-forward \(g\) & Attention \(g\)
    & ratio criterion & gain criterion \\
\midrule
Llama-3.1-8B & \stagxs  & 16 & 16
  & feed-forward, \(T{=}4\) & attention, \(\tau{=}10^{-8}\) \\
 & \stagxxs & 32 & \(v\) 16, \(q,k,o\) 32
  & feed-forward, \(T{=}4\) & attention, \(\tau{=}10^{-9}\) \\
\midrule
Phi-4 & \stagxs  & 16 & 16
  & attention, down, \(T{=}4\) & --- \\
 & \stagxxs & 32 & 32 (fused)
    & down, \(T{=}4\) & attention, \(\tau{=}10^{-8}\) \\
\midrule
OLMo-2-7B & \stagxs & 16 & 16
  & all, \(T{=}8\) & --- \\
 & \stagxxs & 32 & 32
  & feed-forward, \(T{=}4\) & attention, \(\tau{=}10^{-7}\) \\
\bottomrule
\end{tabular}
\end{table}

The remaining settings are shared by all models and configurations.
The gain criterion always scores the activation-weighted two-bit error.
The AWQ scale \(a_c\) of input channel \(c\) is proportional to
\(\mathbb{E}[x_c^2]^{\alpha/2}\), with \(\alpha=0.25\), raised to \(0.5\) on the
down projections.  The
codebooks of each matrix are fitted to an initial pair per group, which
is never stored.  The initial pair places the four levels symmetrically, with a step
of \(0.29\) of the group range and the outer levels \(6\%\) in from
either end.  On the layers sampled in Section~\ref{sec:source}, the
fitted grids also stay narrower than the group range.

\section{Optimality of the Dyadic Schedule}\label{app:dyadic}

We show that the dyadic schedule of
Equation~\eqref{eq:dyadic-schedule} is optimal when
(i) every code cell is an interval, (ii) one refinement bit splits it
into two interval children, (iii) the conditional density inside the
parent cell is uniform, and (iv) each child is reconstructed by one
scalar under squared error.

\paragraph{Proposition.}
Let \(X\) be uniform on \([a,b]\), split at a threshold \(\theta\)
into \([a,\theta)\) and \([\theta,b]\), each reconstructed by one
value.  The minimum mean-squared-error design is the midpoint split,
\begin{equation}
  \theta^\star=\frac{a+b}{2},\qquad
  r_{\mathrm L}^\star=\frac{3a+b}{4},\qquad
  r_{\mathrm R}^\star=\frac{a+3b}{4}.
  \label{eq:midpoint-split}
\end{equation}

\paragraph{Proof.}
For fixed \(\theta\) the best reconstructions are the child centroids,
\(r_{\mathrm L}=(a+\theta)/2\) and \(r_{\mathrm R}=(\theta+b)/2\).
The distortion is then
\begin{equation}
  D(\theta)
  =\frac{(\theta-a)^3+(b-\theta)^3}{12(b-a)},
  \label{eq:split-distortion}
\end{equation}
whose derivative vanishes only at \(\theta=(a+b)/2\).  Because
\(D''=\tfrac12>0\), the midpoint is the unique minimizer.

\paragraph{Recursion.}
A uniform density restricted to either child is again uniform.  The
proposition therefore applies at every plane.  Starting from a base
cell of width \(s\), plane \(k\) moves the reconstruction by
\(\pm t_k\) with \(t_k=s/2^{k-1}\).  At precision \(M\) the
cells have width \(s_M=s/2^{M-2}\).  The recursion thus recovers the
schedule of Equation~\eqref{eq:dyadic-schedule}, which sets the half-step
of Equation~\eqref{eq:main-stream-additive} and the spacing of
Equation~\eqref{eq:main-affine}.

\paragraph{Scope.}
The recursion covers only the refinement planes above the four-level
base, which is fitted in Section~\ref{sec:source}.  Under a nonuniform
conditional density, the optimal threshold and reconstructions move off
the midpoints.

\paragraph{Testing the schedule against the displacement.}
The centroid displacement of Section~\ref{sec:source} shows that condition~(iii) fails.  We
test whether it calls for a different schedule with one format-wide
factor \(\gamma\) on the refinement steps,
\(\alpha_k=\gamma\,s/2^{\,k-1}\), which adds no per-group metadata.
The cost is
\begin{equation}
  C(\gamma)=\frac14\sum_{M=3}^{6}\frac{D_M(\gamma)}{D_M(1)},
  \qquad
  D_M=\mathbb{E}_{x}\bigl\lVert\bigl(W-\widehat W^{(M)}\bigr)x\bigr\rVert^{2},
  \label{eq:gamma-cost}
\end{equation}
where \(D_M\) is the output error of the precision-\(M\) reconstruction under calibration
activations \(x\) drawn from the same pool as the base fit.  In this test
the pre-grid guard uses the ratio criterion (\(T{=}4\)) on every matrix.
The base is held at its \(\gamma=1\) fit under the first term of
Equation~\eqref{eq:ladder-objective} alone, without the floor term
\(\Phi\) that \stagq adds.  We take
the median of \(C(\gamma)\) over \(28\) matrices of Llama-3.1-8B at the
two group sizes this paper evaluates.  At \(g=16\) the median is lowest at
\(\gamma=1\).  At \(g=32\), however, a rescale to \(1.05\) lowers the median by
\(0.6\%\).  The two group sizes disagree on where the optimum lies.  The \(g=32\) basin is also flat, with the median at \(\gamma=1.1\) within \(0.16\%\) of that at \(1.05\).  The released format keeps \(\gamma=1\).

\section{The Centroid Shape on Further Checkpoints}
\label{sec:appendix-models}

\begin{figure}[t]
  \centering
  \includegraphics[width=\linewidth]{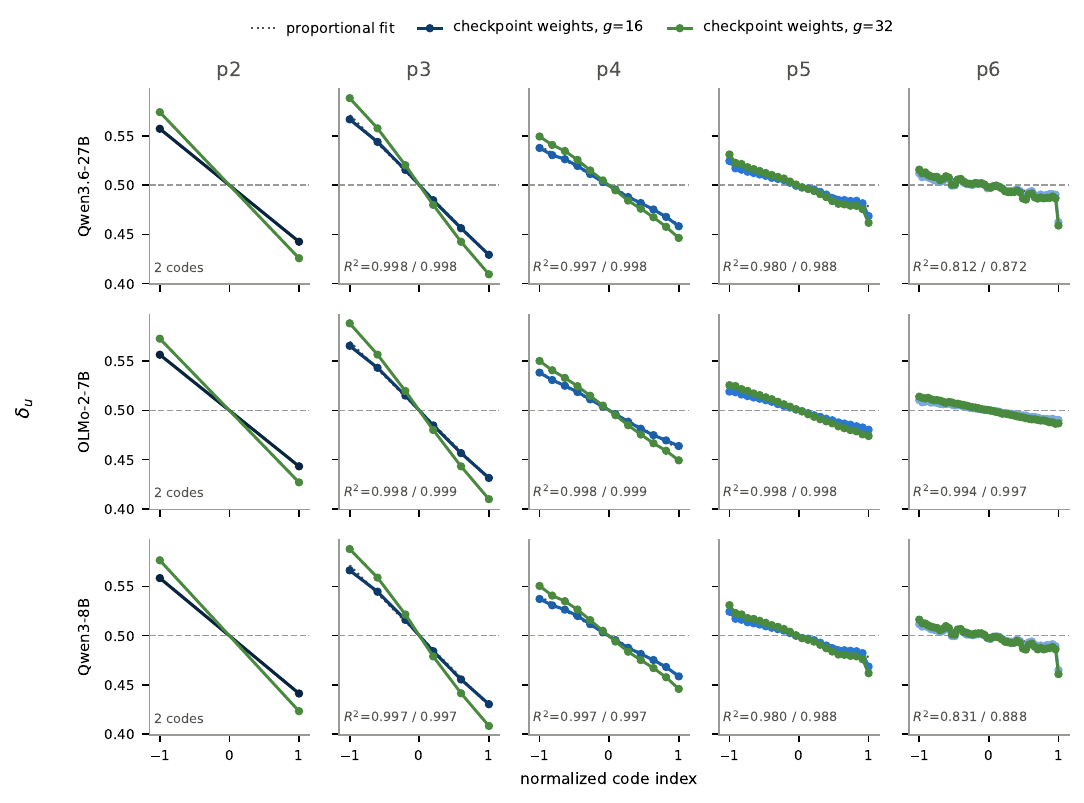}
  \caption{The measurement of Figure~\ref{fig:delta-shape} on the three further
  dense checkpoints (rows: Qwen3.6-27B, OLMo-2-7B, Qwen3-8B) at five
  precisions (columns), on a common
  vertical scale.  Blue: \(g=16\); green: \(g=32\).  \(R^2\) of the proportional fit (blue / green) is
  marked from \(p3\) on.}
  \label{fig:delta-models}
\end{figure}

We repeat the measurement of Section~\ref{sec:source} on three further dense checkpoints, Qwen3.6-27B, OLMo-2-7B~\citep{olmo2} and Qwen3-8B~\citep{qwen3}, at five precisions and both evaluated group sizes (Figure~\ref{fig:delta-models}).  Together with the Llama-3.1-8B checkpoint of Section~\ref{sec:source}, the statistics below cover four checkpoints.  On each checkpoint, including that of Figure~\ref{fig:delta-shape}, we sample \(24\) feed-forward matrices and subsample each to evenly spaced rows, up to four million entries.  We keep only the cells that lie wholly inside the group range, and drop each group's smallest and largest entry.  As a control, we replace every sampled matrix with uniform random values of the same shape.  On this uniform control, \(\delta_u\) lies between \(0.4992\) and \(0.5007\) for every code over the five precisions and both group sizes.

On the three further checkpoints, the straight line of
Section~\ref{sec:source} is reproduced at every group size from \(p3\)
through \(p5\) (Figure~\ref{fig:delta-models}).  The proportional fit is a straight line fitted
to \(\delta_u\) against the code index.  Across the four
checkpoints, its slope at \(p4\) spans \(-0.73\) to \(-0.70\) at
\(g=16\) and \(-0.97\) to \(-0.94\) at \(g=32\), in units of
\(10^{-2}\,\delta_u\) per code step.  From \(p3\) through \(p5\), \(R^2\) lies between
\(0.980\) and \(0.999\), and the crossing at \(\tfrac12\) sits at the
group center.  At these precisions the property Section~\ref{sec:source}
uses, a displacement proportional to the distance from the group center,
therefore holds on each of the four checkpoints.  However, at \(p6\) the
measured curve shifts below \(\tfrac12\) on three of the four
checkpoints, and its crossing moves up to \(0.16\) of the half-width off
the center.  On these three, \(R^2\) drops to \(0.81\)--\(0.83\) at
\(g=16\) and \(0.87\)--\(0.89\) at \(g=32\).  OLMo-2-7B is the exception,
with \(R^2=0.994\) at \(g=16\) and \(0.997\) at \(g=32\).

\section{Encoder Stages}\label{app:encoder}

\paragraph{Scale.}
The encoder calibrates all three reported models on one pool of \(200\) fixed
\(2{,}048\)-token C4~\citep{c4} windows.  The quantizer operates on the
AWQ-scaled matrix~\citep{awq} \(W\operatorname{diag}(a)\), with
\(a_c\propto \mathbb{E}[x_c^2]^{\alpha/2}\) for input channel \(c\) and AWQ exponent \(\alpha\).  The inverse scale folds into a
preceding normalization where one exists.  Where none exists, the
decoder we time applies this inverse scale at run time on Llama-3.1-8B and Phi-4
and omits it on OLMo-2-7B.

\paragraph{Base.}
Each of the two \texttt{fp16} codebooks of Section~\ref{sec:source} is placed by one of three rules.  The rules are uniform placement over the range of the initial pairs (Appendix~\ref{app:configurations}), Lloyd's rule, and Lloyd's rule with its ends anchored at that range.  For each matrix, the encoder picks one of the nine rule combinations by the first term of Equation~\eqref{eq:ladder-objective} alone.  It adds the floor term only in the later
per-group choice among the \(256\) pairs.  However, a matrix whose six-bit error
exceeds a tenth of its two-bit error under the chosen codebooks falls
back to uniform placement.  The base code is nearest rounding, with
no GPTQ error propagation or feedback update.

\paragraph{Guards.}
The pre-grid guard is that of Section~\ref{sec:holdout}, under the
criteria of Table~\ref{tab:configurations}.  On a matrix under the ratio criterion,
and on Phi-4's fused gate/up projection, the post-grid guard takes the
one position per output row with the largest activation-weighted two-bit
error.  Under the gain criterion, it scores each position by that error per stored bit,
takes the three highest-scoring positions per row and keeps those whose score exceeds \(\tau\).
The \(26.9\%\) and \(6.2\%\) of Section~\ref{sec:holdout} are measured
on Phi-4's \(80\) attention matrices at \(g{=}32\), under the gain
criterion at \(\tau{=}10^{-8}\).

\paragraph{Side-record storage.}
Under the gain criterion, a pre-grid entry is an eight-bit symmetric
fixed-point code against one \texttt{fp16} scale per output row.  Its position is
charged at \(\log_2(n/k)+\log_2 e\) bits for a matrix of \(n\) weights
holding out \(k\).  This charge is slightly above the per-entry length of
an enumerative code of the positions.  However, the encoder does not
store such a code.  The layout our kernel reads keeps fixed per-group slots
on attention and per-row slots on the feed-forward matrices instead
(Appendix~\ref{app:limitations}).  Under the ratio criterion, a pre-grid entry is \texttt{fp16}
with a flat index.  A post-grid entry is its \texttt{fp16} residual against the
two-bit base, with a \(\lceil\log_2 c\rceil\)-bit index of its column among the
\(c\) inputs.  Under the gain criterion, each output row also carries a
two-bit count of its post-grid entries.

\paragraph{Group sizes.}
Table~\ref{tab:group-ablation} compares \stagxxs and \stagxs at two bits with the \(g{=}32\)-throughout and \(g{=}16\)-throughout configurations they start from.  At three bits, they gain \(1.59\) and \(0.61\) MMLU points over those configurations, which score \(60.30\) (Table~\ref{tab:ablation}) and \(61.94\).  The second gain lies inside the tie band.
From four bits up, no MMLU or CSR difference from the starting configuration exceeds \(0.42\) points in
magnitude.

\begin{table}[t]
\centering
\caption{Llama-3.1-8B at two bits, with rates in logical \(\bpw\).  The
\(+v\) row changes only the value projection's group size.  The rows marked
\(\to\) change several settings at once.  The \(g{=}32\) row is the
both-guards row of Table~\ref{tab:ablation}.  AnyBCQ scores \(19.03\) and
\(35.33\) at \(2.38~\bpw\).}
\label{tab:group-ablation}
\footnotesize
\begin{tabular}{lrrr}
\toprule
& \bpw & Wiki \(\downarrow\) & MMLU \(\uparrow\) \\
\midrule
\(g{=}32\) throughout          & 2.26 & 20.86 & 25.09 \\
\quad\(+\ v\) at \(g{=}16\)     & 2.26 & 16.68 & 29.46 \\
\quad\(\to\) \stagxxs           & 2.35 & 12.77 & 38.38 \\
\(g{=}16\) throughout          & 2.51 & 10.82 & 43.71 \\
\quad\(\to\) \stagxs            & 2.64 & 9.68 & 48.67 \\
\bottomrule
\end{tabular}
\end{table}

\section{Per-Shape GEMV Latency}\label{app:gemv}

The cuBLAS column of Table~\ref{tab:gemv-fourway} is the median of
three timings per shape, one at each bit width.  Table~\ref{tab:gemv-a100}
divides each kernel's timing by the cuBLAS timing taken at the same
width.  Its speedups can therefore differ slightly from those recomputed
from this column.  On the three attention
shapes, the \stagq layout we time reserves an extra half bit per weight
for per-group guard space.  It reserves none on the six feed-forward
shapes, although real feed-forward matrices also hold guard entries.
The feed-forward timings therefore favor \stagq over the other kernels.
On the attention shapes, the \stagq kernel loses five of the six
comparisons at two bits.  It wins ten of the twelve at three and four
bits.

\begin{table}[t]
\centering
\caption{Batch-one GEMV latency (\(\mu\)s) on one A100 (40 GB) for
the nine shapes (input \(\times\) output) of AnyBCQ's kernel table, from
Llama-3-8B, Phi-4 and Llama-3-70B in row order.  Bold marks the fastest quantized kernel per shape and
width.}
\label{tab:gemv-fourway}
\small
\setlength{\tabcolsep}{3.2pt}
\begin{tabular}{lr@{\hskip 7pt}rrr@{\hskip 7pt}rrr@{\hskip 7pt}rrr}
\toprule
& cuBLAS & \multicolumn{3}{c}{Any-Precision}
& \multicolumn{3}{c}{AnyBCQ} & \multicolumn{3}{c}{\stagq} \\
\cmidrule(lr){2-2}\cmidrule(lr){3-5}\cmidrule(lr){6-8}\cmidrule(lr){9-11}
Shape & 16-bit & 2-bit & 3-bit & 4-bit & 2-bit & 3-bit & 4-bit
      & 2-bit & 3-bit & 4-bit \\
\midrule
8B qkv/o \(4096{\times}4096\) & 29.4 & \textbf{11.5} & \textbf{14.3} & 16.6 & 12.2 & 15.3 & 17.9 & 13.2 & 14.7 & \textbf{14.6} \\
8B down \(14336{\times}4096\) & 97.8 & 29.2 & 38.6 & 47.1 & 25.1 & 32.5 & 40.2 & \textbf{24.1} & \textbf{29.0} & \textbf{32.6} \\
8B gate/up \(4096{\times}14336\) & 97.9 & 27.3 & 34.7 & 41.9 & 24.9 & 32.5 & 40.7 & \textbf{23.7} & \textbf{28.9} & \textbf{32.8} \\
14B qkv/o \(5120{\times}5120\) & 50.1 & 15.5 & 18.7 & 22.5 & \textbf{14.3} & \textbf{18.2} & 22.2 & 16.3 & 18.4 & \textbf{20.0} \\
14B down \(17920{\times}5120\) & 141.3 & 40.8 & 48.4 & 54.4 & 32.1 & 37.7 & 49.4 & \textbf{30.8} & \textbf{36.4} & \textbf{41.6} \\
14B gate/up \(5120{\times}17920\) & 137.5 & 33.7 & 43.6 & 48.7 & 30.2 & 40.1 & 47.8 & \textbf{28.6} & \textbf{37.9} & \textbf{41.5} \\
70B qkv/o \(8192{\times}8192\) & 104.9 & 34.8 & 39.3 & 47.8 & \textbf{26.6} & 35.2 & 43.7 & 28.9 & \textbf{34.3} & \textbf{37.8} \\
70B down \(28672{\times}8192\) & 344.1 & 80.9 & 124.8 & 120.0 & 62.0 & 86.2 & 110.5 & \textbf{58.8} & \textbf{78.7} & \textbf{94.4} \\
70B gate/up \(8192{\times}28672\) & 340.7 & 68.3 & 91.2 & 114.4 & 62.9 & 85.3 & 109.5 & \textbf{59.1} & \textbf{78.8} & \textbf{94.5} \\
\bottomrule
\end{tabular}
\end{table}

\section{Kernel Implementation}\label{app:kernel-impl}

Our kernel is one CUDA C++ GEMV that uses no
CUTLASS, cuBLAS or tensor cores.  It keeps the tiling, half-precision
accumulation and atomic-add structure of the AnyBCQ kernel, so that a timing difference
comes from the format and not from the surrounding code.  A thread block of 256 threads owns a 1024-column
output tile and a 64-row input tile.  In our decode path, each block loads into shared memory a 256-entry
table for each codebook pair its output rows use.  The table maps each
combination of the two codebooks of Section~\ref{sec:source} to
\((lo,s)\).  The lookup runs once per group of \(g\) weights.  The kernel
timed in Table~\ref{tab:gemv-a100} reads this table from constant memory
instead.  The kernel does not read the stored \texttt{fp16} tensors.  A separate
converter builds integer bit planes from the base levels, codebook
indices and codebooks, which the encoder can also write.  The kernel
reads these planes as contiguous \(M\)-bit digits, repacked once at load.
With real weights we ran this path at two bits only.  Both guard stages are applied in the same pass without branching.

\section{The Grid of a Bit-Plane Image Coder}\label{app:jpeg}

JPEG2000 quantizes a wavelet coefficient \(y\) with step \(\Delta\) by the
dead-zone quantizer
\begin{equation}
  q_{\mathrm J}
  =\operatorname{sign}(y)\left\lfloor |y|/\Delta \right\rfloor .
  \label{eq:jpeg-deadzone}
\end{equation}
The magnitude \(|q_{\mathrm J}|\) is written in \(N_b\) bits and coded
one bit-plane at a time, most significant first, with \(b_k\) the bit
at plane \(k\).  Reading \(M\) planes reconstructs a coefficient whose
read bits are not all zero (an all-zero read reconstructs to zero) as
\begin{equation}
  \hat y^{(M)}
  = \operatorname{sign}(y)\,\Delta\!\left(
      \sum_{k=1}^{M} b_k\,2^{\,N_b-k} + \delta\,2^{\,N_b-M}
    \right),
  \label{eq:jpeg-prefix}
\end{equation}
which is Equation~\eqref{eq:direct-sum} under
\(\alpha_k=\operatorname{sign}(y)\,\Delta\,2^{N_b-k}\),
\(\varphi_k(b_k)=b_k\), and
\(c^{(M)}=\operatorname{sign}(y)\,\delta\,\Delta\,2^{N_b-M}\).

The steps \(\alpha_k\) and digit maps \(\varphi_k\) only write the integer
\(|q_{\mathrm J}|\) in binary.  The base term \(c^{(M)}\) places the reconstruction
inside the cell the unread planes leave open, at a fraction
\(\delta\) of its width.  The fraction \(\delta\) is therefore the only
design choice in this grid.  Under the Laplacian model commonly used for wavelet detail
coefficients, the reconstruction that minimizes squared error sits at a
fixed fraction of every cell read at one precision.
\citet{pearlman2008wavelet} report that a single \(\delta\approx0.38\),
measured on the magnitude from the cell edge nearer zero, works well in practice.

\end{document}